\documentclass{SCIS2021}

\begin{document}
\ArticleType{RESEARCH PAPER}
\Year{2020}
\Month{}
\Vol{}
\No{}
\DOI{}
\ArtNo{}
\ReceiveDate{}
\ReviseDate{}
\AcceptDate{}
\OnlineDate{}

\title{Sequential Multi-task Learning with Task Dependency for Appeal Judgment Prediction}{Sequential Multi-task Learning with Task Dependency for Appeal Judgment Prediction}

\author[1]{Lianxin SONG}{}
\author[1]{Xiaohui HAN}{{xiaohhan@gmail.com}}
\author[1]{Guangqi LIU}{}
\author[2]{Wentong WANG}{}
\author[3]{\\Chaoran CUI}{}
\author[2]{Yilong YIN}{}

\AuthorMark{Song L X}

\AuthorCitation{Song L X, Han X H, Liu G Q, et al}


\address[1]{Shandong Computer Science Center (National Supercomputer Center in Jinan), Shandong Provincial Key Laboratory \\of Computer Networks, Qilu University of Technology (Shandong Academy of Sciences), Jinan {\rm 250014},  China}
\address[2]{School of Software, Shandong University, Jinan {\rm 250000}, China}
\address[3]{School of Computer Science and Technology, Shandong University of Finance and Economics, Jinan {\rm 250014}, China}

\abstract{Legal Judgment Prediction (LJP) aims to automatically predict judgment results, such as charges, relevant law articles, and the term of penalty. It plays a vital role in legal assistant systems and has become a popular research topic in recent years. This paper concerns a worthwhile but not well-studied LJP task, Appeal judgment Prediction (AJP), which predicts the judgment of an appellate court on an appeal case based on the textual description of case facts and grounds of appeal. There are two significant challenges in practice to solve the AJP task. One is how to model the appeal judgment procedure appropriately. The other is how to improve the interpretability of the prediction results. We propose a \textit{Sequential Multi-task Learning Framework with Task Dependency for Appeal Judgement Prediction} (SMAJudge) to address these challenges. SMAJudge utilizes two sequential components to model the complete proceeding from the lower court to the appellate court and employs an attention mechanism to make the prediction more explainable, which handles the challenges of AJP effectively. Experimental results obtained with a dataset consisting of more than 30K appeal judgment documents have revealed the effectiveness and superiority of SMAJudge.}


\maketitle

\section{Introduction}
With the development of artificial intelligence, Legal Judgment Prediction (LJP) has become a popular research topic in recent years. LJP aims to automatically predict judgment results from the description of facts of a specific case. It plays a vital role in legal assistant systems, benefiting legal professionals (e.g., judges and lawyers) and ordinary people. On the one hand, such systems can provide a handy reference for professionals and improve their work efficiency. On the other hand, they can supply ordinary people unfamiliar with legal terminology and complex procedures with legal consulting.

LJP has been studied for decades, and most existing works regard this task as a text classification problem. Researchers have proposed various methods based on machine learning and deep learning models and made significant progress in LJP tasks like  like predicting charges \cite{1}, relevant law articles \cite{2}, and the term of penalty \cite{3}. Some of the existing works have already achieved an accuracy of over 90\% on predicting charges and relevant law articles, which is very close to human judges.

This paper concerns another worthwhile but not well-studied LJP task, i.e., Appeal judgment Prediction (AJP). 
In law, an appeal is the legal proceeding by which a case is brought before an appellate court to review a lower court's judgment. 
A party to the case unsatisfied with the lower court's decision might be able to file a petition to the appellate court and challenge that decision on specific grounds.  
Grounds of appeal could include errors of law, fact, procedure, or due process. It may also be possible to appeal based on new evidence. 
The appellate court generally affirms, reverses, or remands a lower court's judgment, depending on whether the judgment was legally sound or not. Accordingly, the goal of the AJP task is to predict the ruling of an appellate court over the lower court's judgment.

AJP is also very helpful for building legal assistant systems. For instance, it can help appellants estimate their appeal interest, thus helping to curb frivolous appeals. It also can assist lower courts in checking the defects of their decisions in advance.  
However, only limited research has paid attention to AJP. In \cite{4}, Katz et al. proposed a supervised learning approach to predict appeal decisions of the U.S. supreme court. However, their approach needs complex manual feature engineering. Moreover, their approach leveraged many features designed towards the U.S. legal system, such as historical behavior features of justices and courts, which are difficult to be extended to different law systems.

This paper focuses on conducting AJP only with the textual description of case facts and grounds of appeal as input, which has better universality for different legal systems. The task is not a trivial problem, and there are two major challenges in practice.
\begin{itemize}
	\item \emph{Challenge I: How to appropriately model an appellate court's decision process?} An appellate court typically makes its decision by reviewing a lower court's trial proceedings. In addition, the grounds of appeal also have critical impacts on the appellate court's decision. We should take all these matters into account when building the prediction model. Existing single-trial LJP methods can not be directly used to solve the AJP task.
	\item \emph{Challenge II: How to improve the interpretability of the prediction results?} In practice, the appellate court typically issues its judgment on an appeal case accompanied by an opinion explaining the rationale. However, the interpretability of results of most machine learning models is weak due to their ``black-box'' characteristics. Only giving the prediction results is not enough to convince people, especially legal professionals. 
\end{itemize}

To solve the AJP task and address the above two challenges, we propose a \textbf{S}equential \textbf{M}ulti-task Learning Framework with Task Dependency for \textbf{A}ppeal \textbf{Judge}ment Prediction (SMAJudge). 
SMAJudge employs two multi-task learning components with dependency among subtasks to model the proceedings in a lower court and an appellate court, respectively. The two components are connected sequentially to model the complete proceeding from the lower court to the appellate court. The prediction of the appeal result is formulated as a subtask in the component corresponding to the appellate court and learned jointly with other subtasks. SMAJudge also utilizes an attention mechanism to improve the interpretability of the prediction results. We evaluate SMAJudge with a dataset composed of 33238 real-word appeal judgment documents. The experimental results demonstrate the superiority of SMAJudge over competitive baselines.

To summarize, the key contributions of this paper are as follows.
\begin{itemize}
	\item We propose SMAJudge, which predicts the judgment result of an appeal case in an end-to-end manner. Besides, SMAJudge only requires the description of case facts and grounds of appeal as input. Neither manual feature engineering nor external auxiliary data beyond the appeal case are needed. Compared with the limited existing AJP research, SMAJudge is easier to be extended to other legal systems
	
	\item In SMAJudge, we propose a sequential multi-task learning framework with task dependency. Our experimental results reveal that the proposed framework provides an effective solution to Challenge I of the AJP task.
	\item We also propose an attention mechanism to capture the impact of grounds of appeal on the appeal judgment. Our experimental results demonstrate that the proposed attention mechanism effectively solves both Challenge I and Challenge II of the AJP task.	
\end{itemize}

The rest of this paper is organized as follows. Section \ref{sec:relatedwork} gives a short review of existing LJP techniques. Section \ref{sec:statement} formally defines the AJP task. Section \ref{sec:framework} proposes the SMAJudge framework. Section \ref{sec:experiments} reports and analyzes the experimental results. Section \ref{sec:conclusions} concludes this paper and offers possible future work directions.

\section{Related Work}
\label{sec:relatedwork}

LJP research can be traced back to the 1960s.  Early works usually explore mathematical and statistical algorithms to analyze existing legal cases in specific scenarios and succeed with small-scale datasets \cite{5,6,7,8,9,10}. Some later works designed various rule-based expert systems\cite{11,12}, in which the quality of rules heavily depends on human experts' understanding of the law. As the number of rules increases, there might be severe conflicts between rules.

With the development of machine learning and Natural Language Processing (NLP) techniques, more researchers addressed LJP under text classification frameworks. At first, most studies attempt to extract shallow text features from case descriptions or annotations (e.g., locations, terms, and dates) in textual form. Then these studies utilize classification models (e.g., SVM and Random Forest) to predict judgment results based on the extracted features\cite{13}. For instance, Lin et al. \cite{14} defined 20 legal terms as features to predict the judgments of robbery and intimidation cases. Aletras et al. \cite{15} built an SVM classifier with n-gram and topic features to predict the European Court of Human Rights decisions. Sulea et al. proposed a similar technique in \cite{16}. However, due to the limited semantic representation ability of shallow textual features and manually designed factors, the performance and generalization ability of these methods are usually weak.

In recent years, deep learning models have been successfully applied in many NLP tasks\cite{17,18,19}. Motivated by this, researchers began to handle LJP with deep learning models and obtain better performance than conventional machine learning methods. For example, Luo et al. \cite{1} proposed an attention-based neural network to capture the correlation between case facts and related law articles to promote charge prediction. Wang et al. \cite{2} proposed a hierarchical deep network with co-attention to predict relevant law articles based on case facts. 

Some studies integrate multiple LJP tasks into a multi-task learning framework to improve the performance of each one by information transfer among them. Hu et al. \cite{20} incorporate the charge prediction task with the inference of several manually selected legal attributes to better predict few-shot and confusing charges. Li et al. \cite{21} proposed a multi-channel attention neural network to jointly predict charges, relevant law articles, and the term of penalty. Xu et al. \cite{3} proposed LADAN, which also addressed the above three tasks jointly by multi-task learning. LADAN can effectively distinguish subtle differences between confusing law articles with a community-based graph neural network. Zhong et al. \cite{22} proposed TOPJUDGE, which formalizes the dependencies among subtasks as a DAG (Directed Acyclic Graph) and utilizes a topological multi-task learning framework for effectively handling these subtasks jointly. Following their work, Yang et al. \cite{23} further considered the backward dependencies between the prediction results of subtasks and proposed a refined framework with a backward verification mechanism. Ma et al. \cite{24} utilized multi-stage judicial data, e.g., court debates, facts, and claims, to jointly learn the identification of the legal facts in the court debate and predict the judgment results of each claim.

Although researchers have made significant progress on LJP, the task of AJP has not been well studied in the literature. Katz et al. \cite{4} proposed a random-tree-based model to predict the U.S. Supreme Court decisions on appeal cases. They derived case features by manual feature engineering from  SCDB (Supreme Court Database) data. However, many features are specifically designed for the U.S. legal system, such as the behavior features of justices and courts. Computing these features requires accumulating external auxiliary data beyond a particular case (e.g., historical behavior data about justices and courts). Unlike \cite{4}, SMAJudge conducts AJP only with the textual description of case facts and grounds of appeal as input. No additional external data is required. Moreover, SMAJudge predicts the judgment result from the input in an end-to-end manner without feature engineering. Therefore, SMAJudge is more accessible to other law systems.

\begin{figure*}[htbp]
\centerline{\includegraphics[width=15cm]{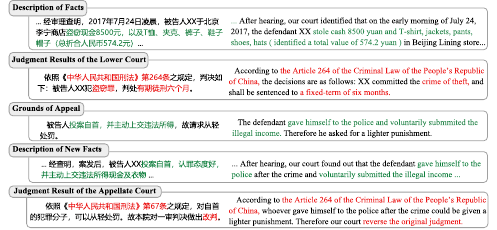}}
\caption{An example of an appeal judgment document}
\label{fig:lawtext}
\end{figure*}

\section{Problem Formulation}
\label{sec:statement}
Following existing LJP studies, we treat AJP as a textual classification problem, which takes the textual description of an appeal case as the input and outputs the judgment of the appellate court. The textual input includes the facts of the case, the decisions of the lower court, and the appeal grounds. We use a corpus of judgment documents of finished appeal cases as the training data to learn an automatic appeal judgment predictor.

As shown in Fig.\ref{fig:lawtext},an appeal judgment document $d$ of a specific appeal case is formally represented as:
\begin{equation}
d = [{f^l},{r^l},{g},{f^{a}},{r^{a}}],
\label{eq}
\end{equation}
where $f^l$ denotes the description of facts confirmed by the lower court, $r^l$ denotes the judgment results of a lower court, $g$ denotes the description of grounds for appeal, $f^a$ denotes the description of new facts or evidence developing after the trial in the lower court, and $r^a$ denotes the judgment result of an appellate court. 

Specifically, we only focus on criminal appeal cases in this paper. The judgment result of a criminal case typically consists of several detailed aspects. Here, we take three aspects  into account, i.e., relevant law articles, charges, and the term of penalty. 
Thus, $r^l$ can be represented as a triple:
\begin{equation}
	r^l=[y^{ll}, y^{lc}, y^{lp}],
\end{equation}
where $y^{ll}$, $y^{lc}$, and $y^{lp}$ denote relevant law articles, charges, and the term of penalty determined by the lower court, respectively. For the appeal procedure, we consider two aspects of the judgment result. One is the appellate court's ruling over the lower court's judgment result, $y^{ar}$. The other is the legal basis (i.e., relevant law articles) for the appellate court's ruling, $y^{al}$. Hence, $r^a$ can be represented as:
\begin{equation}
	r^a=[y^{ar}, y^{al}].
\end{equation}

Based on the representation of an appeal judgment document, we formally define the AJP task as follows.

\textbf{Definition 1. Appeal Judgment Prediction Task. }\textit{Given a corpus of $N$ appeal judgment documents $D=\{d_i\}^N_{i=1}$, the appeal judgment prediction task aims at learning a prediction model $pre()$ with $D$. When the textual description of a new appeal case $d_{k}=[f^l_k,r^l_k,g_k,f^a_k]$ is provided, $pre()$ can predict the judgment result ${\hat{r}}^a_k$ of the appellate court with $d_k$ as input, i.e.,}
\begin{equation}
	{\hat{r}}^a_k=pre(d_k)=pre([f^l_k,r^l_k,g_k,f^a_k]).
\end{equation}

Note that there could be complex outcomes of the appellate court's ruling over the lower court's judgment. For example, the appellate court might affirm some of the lower court's rulings and reverse others. However, to make the problem simple, we only predict whether the appellate court will affirm the lower court's whole judgment or not, i.e.,
\begin{equation}
y^{ar}=
\begin{cases}
0, & \textit{if the appellate court affirms all the lower}\\
    & \textit{court's rulings,}\\
1, & \textit{otherwise.}
\end{cases}
\end{equation}
Moreover, we only focus on cases with one defendant.

\begin{figure*}[htbp]
\centerline{\includegraphics[width=15cm]{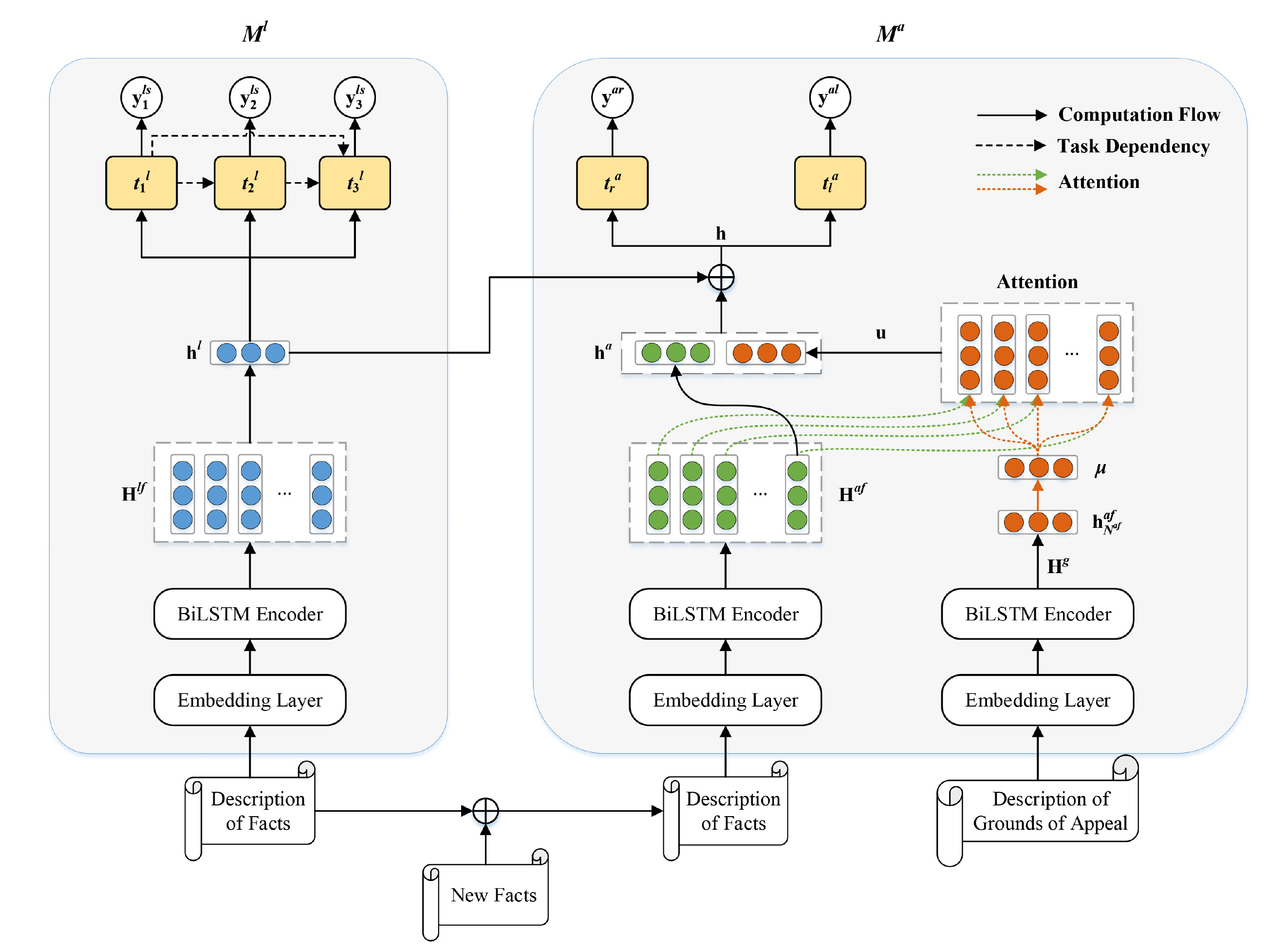}}
\caption{Architecture of SMAJudge}
\label{fig:framework}
\end{figure*}

\section{Proposed Framework}
\label{sec:framework}
\subsection{Overview of SMAJudge}
As shown in Fig.\ref{fig:framework}, SMAJudge consists of two sequential multi-task learning components, $M^l$ and $M^a$, which model the proceedings in the lower court and the appellate court, respectively. The architectures of the two components are as follows. 
\begin{itemize}
	\item \textbf{$M^l$} encodes the description of facts confirmed by the lower court into a real-value vector and uses it to solve three subtasks, i.e., predicting relevant law articles, charges, and the term of penalty. There are explicit dependencies among these subtasks, which simulate a human judge's decision logic.
	\item \textbf{$M^a$} takes the descriptions of facts confirmed by the lower court, possible newly developed facts, and appeal grounds as input and encodes them into a real-value vector with an attention mechanism. The vector is combined with the one encoded in $M^l$ to solve two subtasks, i.e., predicting the appellate court's ruling on the lower court's judgment and relevant law articles.
\end{itemize}

In the rest of Section \ref{sec:framework}, we first present the text encoding strategies of the two components and then give the details of the subtasks in them. Table \ref{tab:symbols} gives the symbols used in this paper. 

\begin{table}[htbp]
\footnotesize
\caption{Symbols used in the paper}
\begin{center}
\label{tab:symbols}
\tabcolsep 15pt 
\begin{tabular}{ll}
\toprule
  \textbf{Symbol}&\textbf{Description}  \\\hline
$D$&corpus of appeal judgment documents\\
$N$&number of appeal judgment documents in $D$\\
$d_i$&$i$th appeal judgment document in $D$\\
$f^*$\footnote&textual description of case facts\\
$r^*$&judgment results\\
$g$&textual description of grounds of appeal\\
$\mathbf{y}^*$/$\hat{\mathbf{y}}^*$&ground-truth/predicted distribution of a subtask\\
$M^*$&multi-task learning component in SMAJudge\\
$w^*$&word in textual descriptions\\
$\mathbf{w}^*$&embedding vector of a word\\
$N^*$&number of words in the textual description of facts or grounds of appeal\\
$\mathbf{h}^*$&hidden state in SMAJudge\\
$\mathbf{H}^{*}$&sequence of hidden states\\
$\mathbf{W}^*$,$\mathbf{b}^*$&weight matrix and bias in SMAJudge\\
$\alpha_i$&attention value\\
$\mathbf{u}$&encoding vector of facts impacted by the grounds of appeal\\
$\mathbf{z}^*$&value of gates in an LSTM\\
$\mathbf{C}^*$&value of the memory cell in an LSTM\\
$t^*$& subtaks in SMAJudge\\
$T$&ordered list of subtasks\\
$\mathcal{L}^*$&loss function of a subtask\\
$\mathcal{L}$&overall loss of SMAJudge\\
\bottomrule
\end{tabular}
\end{center}
\end{table}

\subsection{Text Encoding}

Various deep learning models can be utilized to encode textual content into a real-value vector, such as LSTM, GRU, CNN, BERT (Bidirectional Encoder Representations from Transformers), and GTP (Generative Pre-Training). Although large-scale pre-training models (e.g., BERT and GTP) have shown outstanding ability in many NLP scenarios, some studies claim they do not work very well in legal text processing tasks\cite{25}. Moreover, the resource and time cost to fine-tuning BERT or GPT on long legal text is very high. Therefore, we employ a Bidirectional-LSTM (BiLSTM) as the text encoder in SMAJudge. BiLSTM has been widely used and performs well in LJP research. It consists of two LSTM functions, which read the input sequence forward and backward. We refer to the two functions as $\overrightarrow{\text{LSTM}(\cdot)}$ and $\overleftarrow{\text{LSTM}(\cdot)}$. Note that other models can also work as the encoder. However, the discussion on choosing encoders is out of our primary focus.

\footnotetext[1]{``*'' is a wildcard that refers to any superscript or subscript.}

\subsubsection{Text Encoding in $M^l$} The input of $M^l$ is the description of facts confirmed by the lower court, i.e., $f^l$. Let $w^{lf}$=$\{w^{lf}_i\}^{N^{lf}}_{i=1}$ be the sequence of words in $f^l$, where $w^{lf}_i$ and $N^{lf}$ are the $i$th word and the total number of words in $f^l$, respectively. We first embed each word $w^{lf}_i$  into a $k$-dimensional vector using the word2vec model. Let $\mathbf{w}^{lf}=\{\mathbf{w}^{lf}_i\}^{N^{lf}}_{i=1}$ denote the word embedding sequence of $w^{lf}$, where $\mathbf{w}^{lf}_i$ is the embedding vector of $w^{lf}_i$. Then, a BiLSTM encodes $\mathbf{w}^{lf}$ into a hidden state sequence $\mathbf{H}^{lf}= \{\mathbf{h}_i^{lf}\}_{i=1}^{N^{lf}}$, where
\begin{equation}
	\mathbf{h}^{lf}_i = [\overrightarrow{\text{LSTM}^{l}(\mathbf{w}^{lf}_i)},\overleftarrow{\text{LSTM}^{l}(\mathbf{w}^{lf}_i)}].
\end{equation}

We use the last hidden state $\mathbf{h}^{lf}_{N^{lf}}$ in $\mathbf{H}^{lf}$ to represent the lower court's understanding of the facts and refer to it with a simpler notation $\mathbf{h}^l$ hereafter.

\subsubsection{Text Encoding in $M^a$} When reviewing the judgment of a lower court, an appellate court will focus on the parts of facts related to the grounds of appeal. To capture the impact of grounds of appeal on the appellate court's understanding of case facts, we design an attention mechanism to make the encoding procedure in $M^a$ pay more attention to the parts having solid semantic relationships with the appeal grounds in the description of facts. Let $\mathbf{w}^{af}=\{\mathbf{w}^{af}_{i}\}^{N^{af}}_{i=1}$ and $\mathbf{w}^{g}=\{\mathbf{w}^{g}_{i}\}^{N^{g}}_{i=1}$ denote the sequences of word embedding vectors of $f^a$ and $g$, where $N^{af}$ and $N^{g}$ are the number of words in $f^a$ and $g$, respectively. We employ another two BiLSTMs to encode $\mathbf{w}^{af}$ and $\mathbf{w}^{g}$, respectively. After encoding, we can obtain the corresponding sequences of hidden states ${\mathbf{H}^{af}} = \{\mathbf{h}_i^{af}\}_{i = 1}^{N^{af}}$ and ${\mathbf{H}^{g}} = \{\mathbf{h}_i^{g}\}_{i = 1}^{N^{g}}$, where
\begin{equation}
	\mathbf{h}^{af}_i = [\overrightarrow{\text{LSTM}^{a}(\mathbf{w}^{af}_i)},\overleftarrow{\text{LSTM}^{a}(\mathbf{w}^{af}_i)}],
\end{equation}
\begin{equation}
	\mathbf{h}^{g}_i = [\overrightarrow{\text{LSTM}^{g}(\mathbf{w}^{g}_i)},\overleftarrow{\text{LSTM}^{g}(\mathbf{w}^{g}_i)}].
\end{equation}

Based on $\mathbf{h}_{N_g}^{g}$, the last hidden state in $\mathbf{H}^{g}$, we generate a context vector $\bm{\mu}$ for the description of facts as:
\begin{equation}
	\bm{\mu} = {\mathbf{W}^g}{\mathbf{h}^{g}_{N^g}} + {\mathbf{b}^g},
\end{equation}
where ${\mathbf{W}^{g}}$ is the weight matrix, and $\mathbf{b}^g$ is the bias. Then, the encoding procedure of $M^a$ continues to compute a sequence of attention values $[{\alpha _1},{\alpha _2}, \cdots ,{\alpha _{N^ g}}]$ satisfying $\alpha_i\in[0,1]$ and $\sum_i^{N^g} \alpha_i=1$ as:
\begin{equation}
{\alpha _i} = \frac{{\exp (\tanh {{({\mathbf{W}^{af}}\mathbf{h}_i^{af})}^T}{\bm{\mu}})}}{{\sum\nolimits _j {\exp (\tanh {{({\mathbf{W}^{af}}\mathbf{h}_j^{af})}^T}{\bm{\mu}})}}}, 
\end{equation}
where $\mathbf{W}^{af}$ is the weight matrix. The encoding vector of facts impacted by the grounds of appeal is computed as:
\begin{equation}
\mathbf{u} = \sum\limits_{i = 1}^{N^{af}} {{\alpha _i}\mathbf{h}_i^{af}}.
\end{equation}

Finally, we concatenate $\mathbf{u}$ and $\mathbf{h}^{af}_{N^{af}}$ into a vector $\mathbf{h}^a$ to represent the appellate court's understanding of case facts, i.e.,
\begin{equation}
	\mathbf{h}^a=\mathbf{h}^{af}_{N^{af}}\oplus\mathbf{u},
\end{equation}
where $\oplus$ denotes a concatenation operation.

\subsection{Prediction Subtasks in $M^l$}
\label{sec:jpinml}
As mentioned in Section \ref{sec:statement}, a human judge typically determines detailed aspects of a judgment in a particular order. For instance, a judge in a civil law system determines the relevant law articles based on case facts at first and then confirms the charges according to the instructions of relevant law articles. Finally, the judge decides the term of penalty based on relevant law articles and charges. In \cite{22}, Zhong et al. formalize the determination of judgment aspects as subtasks in a multi-task learning framework and represent the dependencies among subtasks with a Directed Acyclic Graph. Inspired by their work, in $M^l$, we adopt a multi-task learning framework with dependencies among subtasks to simulate the decision process in a lower court.

Let $t^l_1$, $t^l_2$, and $t^l_3$ refer to the prediction tasks of relevant law articles, charges, and the term of penalty, respectively. We use $t^l_i\lhd t^l_j$ to denote the $t^l_j$'s dependency on $t^l_i$. All the subtasks that $t^l_j$ depends on constitute a dependency set $D^l_j=\{t^l_i|t^l_i\lhd t^l_j\}$. Our experimental dataset is composed of appeal judgment documents produced by the Chinese legal system, which is an instance of the civil law system. Therefore, we define the dependency among subtasks with an ordered list $T^l=[t^l_1,t^l_2,t^l_3]$ satisfying the following constraint:

\begin{equation}
	i<j, \forall (i,j)\in\{(i,j)|t^l_i\in D^l_j\}.
\end{equation}

\begin{figure}[htbp]
\centerline{\includegraphics[width=6cm]{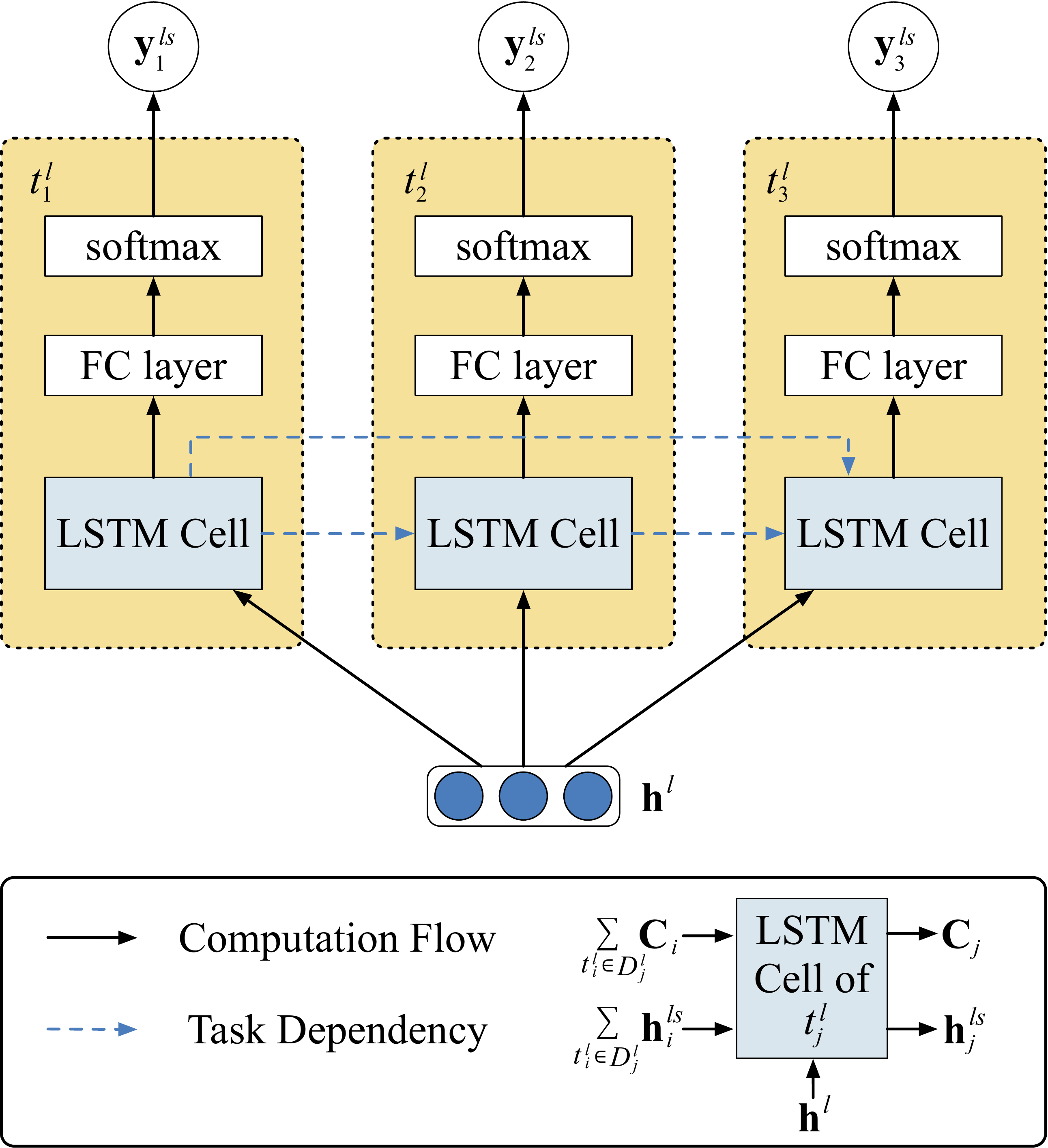}}
\caption{Dependencies among subtasks in $M^l$}
\label{fig:topological}
\end{figure}

Figure \ref{fig:topological} illustrates the dependencies among subtasks defined by $T^l$.  Note that one can easily redefine the subtasks and dependencies to make the model fit other legal systems.

We design a special LSTM to predict the results of subtasks according to the dependency order defined by $T^l$. As depicted in Figure \ref{fig:topological}, each cell in the LSTM corresponds to a subtask $t^l_i\in T^l$. For $t^l_i$, its target value $y^l_i$ is predicted based on the encoding vector $\mathbf{h}^l$ and the results of its dependent subtasks. Precisely, to predict $y^l_i$, we first compute

\begin{equation}
	\mathbf{z}^{F}_j=\sigma\Bigg (\sum\limits_{{t^l_i} \in {D^l_j}} {({\mathbf{W}^{F}_{i,j}}\left[{\begin{array}{*{20}{c}}
{{\mathbf{h}^{ls}_i}}\\
{{\mathbf{h}^l}}
\end{array}} \right])}  + {\mathbf{b}^{F}_j}\Bigg ),
\end{equation}

\begin{equation}
	\mathbf{z}^{I}_j=\sigma\Bigg (\sum\limits_{{t^l_i} \in {D^l_j}} {({\mathbf{W}^{I}_{i,j}}\left[{\begin{array}{*{20}{c}}
{{\mathbf{h}^{ls}_i}}\\
{{\mathbf{h}^l}}
\end{array}} \right])}  + {\mathbf{b}^{I}_j}\Bigg ),
\end{equation}

\begin{equation}
	\mathbf{z}^{O}_j=\sigma\Bigg (\sum\limits_{{t^l_i} \in {D^l_j}} {({\mathbf{W}^{O}_{i,j}}\left[{\begin{array}{*{20}{c}}
{{\mathbf{h}^{ls}_i}}\\
{{\mathbf{h}^l}}
\end{array}} \right])}  + {\mathbf{b}^{O}_j}\Bigg ),
\end{equation}

\begin{equation}
	 \widetilde{\mathbf{C}}_j=\tanh\Bigg (\sum\limits_{{t^l_i} \in {D^l_j}} {({\mathbf{W}^{C}_{i,j}}\left[{\begin{array}{*{20}{c}}
{{\mathbf{h}^{ls}_i}}\\
{{\mathbf{h}^l}}
\end{array}} \right])}  + {\mathbf{b}^{C}_j}\Bigg ),
\end{equation}

\begin{equation}
	\mathbf{C}_j=\mathbf{z}^{F}_j\odot \sum\limits_{{t^l_i} \in {D^l_j}}\mathbf{C}_{i}+\mathbf{z}^{I}_j\odot \widetilde{\mathbf{C}}_j,
\end{equation}

\begin{equation}
	\mathbf{h}^{ls}_j=\mathbf{z}^{O}_j\odot \tanh(\mathbf{C}_j),
\end{equation}
where $\mathbf{z}^F_j$, $\mathbf{z}^I_j$, $\mathbf{z}^O_j$, and $\mathbf{C}_j$ are the outputs of the forget gate, the input gate, the output gate, and the memory cell in the LSTM cell corresponding to $t^l_j$, respectively; $\mathbf{h}^{ls}_i$ and $\mathbf{h}^{ls}_j$ are hidden states of $t^l_i$ and $t^l_j$, respectively; $\mathbf{W}^{F}_{i,j}$, $\mathbf{W}^{I}_{i,j}$, $\mathbf{W}^{O}_{i,j}$, and $\mathbf{W}^{C}_{i,j}$ are weight matrices specific to $t^l_i$ and $t^l_j$; $\mathbf{b}^{F}_j$, $\mathbf{b}^{I}_j$, $\mathbf{b}^{O}_j$, and $\mathbf{b}^{C}_j$ are bias vectors; and $\sigma$ and $\odot$ refer to the sigmoid function and the element-wise product, respectively. 

We further take the hidden state ${\mathbf{h}^{ls}_j}$ as the task-specific input of $t^l_j$ and apply a fully-connected (FC) layer followed by a softmax layer to get the prediction result of $t^l_j$ as:
\begin{equation}
{\hat{\textbf{y}}^{ls}_j} = \text{softmax} (\mathbf{W}_j^{ls}{\mathbf{h}^{ls}_j} + \mathbf{b}_j^{ls}), \label{eq}
\end{equation}
where $\mathbf{W}_j^{ls}$ and $\mathbf{b}_j^{ls}$ are the weight matrix and the bias specific to $t^l_j$, respectively. Finally, we use the cross-entropy between $\hat{\textbf{y}}^{ls}_j$ and the ground-truth $\textbf{y}^{ls}_j$ as the prediction loss of $t^l_j$, which is computed as

\begin{equation}
\mathcal{L}^l_j =  - \sum\limits_{k = 1}^{|Y^l_j|} {\textbf{y}^{ls}_{jk}} \log(\hat{\textbf{y}}^{ls}_{jk}),
\end{equation}
where $Y^l_j$ is a label set specific to $t^l_j$, and $\textbf{y}^{ls}_{jk}$ and $\hat{\textbf{y}}^{ls}_{jk}$ are the $k$th components of $\textbf{y}^{ls}_{j}$ and $\hat{\textbf{y}}^{ls}_{j}$, respectively. For subtasks $t^l_1$ and $t^l_2$, $Y^l_1$ and $Y^l_2$ refer to the set of law articles and charges, respectively. For  subtask $t^l_3$, we divide the terms of penalty into 11 non-overlapping intervals as listed in Table \ref{tab:intervals}, and $Y^l_3$ refers to the set of these 11 intervals.

\begin{table}[htbp]
\footnotesize
\caption{Intervals of the Term of Penalty}
\begin{center}
\label{tab:intervals}
\tabcolsep 25pt 
\begin{tabular}{cc}
\toprule
  Interval Index & Length of the Term of Penalty  \\\hline
 1&No penalty \\
  2& 0$\sim$6 months \\
  3&7$\sim$9 months \\
  4&10$\sim$12 months \\
  5 &1$\sim$2 years \\
  6 &3$\sim$4 years \\
  7 &5$\sim$6 years \\
  8 &7$\sim$8 years \\
  9 &9$\sim$10 years \\
  10 &More than 10years \\
  11 &Death or Life Imprisonment \\
\bottomrule
\end{tabular}
\end{center}
\end{table}

\subsection{Prediction Subtasks in $M^a$}

In $M^a$, there are two subtasks, i.e., predicting the appellate court's ruling over the lower court's judgment (i.e., $y^{ar}$) and the relevant law articles (i.e., $y^{al}$). We refer to the two subtasks as $t^a_r$ and $t^a_l$, respectively. The input of both subtasks is a concatenation vector ${\mathbf{h}} = \mathbf{h}^l\oplus\mathbf{h}^a$. 

To predict $y^{ar}$, we feed $\mathbf{h}$ into a fully-connected layer followed by a sigmoid layer to get the prediction result as:
\begin{equation}
{\hat{y}^{ar}}= \text{sigmoid}(\mathbf{W}^{ar}\mathbf{h} + \mathbf{b}^{ar}),
\end{equation}
where $\mathbf{W}^{ar}$ and $\mathbf{b}^{ar}$ are the weight matrix and bias specific to $t^a_r$, respectively. We compute the corresponding loss as:
\begin{equation}
	\mathcal{L}^a_r = -[y^{ar}\cdot\log(\hat{y}^{ar})+(1-y^{ar})\cdot\log(1-\hat{y}^{ar})],
\end{equation}
where $y^{ar}\in\{0,1\}$ is the ground-truth label.

To predict $y^{al}$, we feed $\mathbf{h}$ into a fully-connected layer followed by a softmax layer as:

\begin{equation}
\mathbf{\hat{y}}^{al} = \text{softmax}({\mathbf{W}^{al}}{\mathbf{h}} + {\mathbf{b}^{al}}),
\end{equation}
where ${\mathbf{W}^{al}}$ is the weight matrix, and ${\mathbf{b}^{al}}$ is the bias. We compute the loss specific to $t^a_l$ as:
\begin{equation}
\mathcal{L}^{a}_{l}=  - \sum\limits_{k = 1}^{|Y^a_l|} {\textbf{y}^{al}_{k}} \log(\hat{\textbf{y}}^{al}_{k}),
\end{equation}
where $Y^a_l$ is the set of law articles specific to $t^a_l$, and $\hat{\textbf{y}}^{al}_{k}$ and $\textbf{y}^{al}_{k}$ are the $k$th components of $\hat{\textbf{y}}^{al}$ and the ground-truth $\textbf{y}^{al}$, respectively.

\subsection{Training and Prediction}
The training objective of SMAJudge is to minimize the integration of losses of all the five subtasks in $M^l$ and $M^a$:

\begin{equation}
\mathcal{L} = {\lambda_1}{\mathcal{L}^l_1} + {\lambda_2}{\mathcal{L}^l_2} + {\lambda_3}{\mathcal{L}^l_3} + {\lambda_4}{\mathcal{L}^a_{r}} + {\lambda_5}{\mathcal{L}^a_{l}},
\end{equation}
where ${\lambda_i}$ are hyper-parameters controlling the weight of the five parts in the integrated loss. We set all ${\lambda_i}$ equally in our experiments, assuming that the five subtasks are of equal importance. We employed the Adam optimization algorithm and the dropout strategy during the training process.

After training, SMAJudge can predict the judgment result of a new appeal case $d_k$ based on its textual description $[f^l_k,r^l_k,g_k,f^a_k]$. SMAJudge first uses $f^l_k$ as the input of $M^l$ and $r^l_k=[y^{ll}_k,y^{lc}_k,y^{lp}_k]$ as the ground-truth of $T=[t^l_1,t^l_2,t^l_3]$ to perform a single round of parameter optimization of $M^l$. The purpose is to fine-tune the parameters in $M^l$ according to the lower court's trial record of $d_k$. Then, SMAJudge uses $[f^l_k,f^a_k,g_k]$ and the fine-tuned parameters to compute the two hidden vectors, $\mathbf{h}^l$ and $\mathbf{h}^a$. Finally, SMAJudge predicts the judgment result of the appellate court $r^a_k=[y^{ar}_{k},y^{al}_{k}]$ with $\mathbf{h}=\mathbf{h}^l\oplus\mathbf{h}^a$.

\section{Experiments and Analysis}
\label{sec:experiments}
In this section, we present an extensive set of experiments to evaluate the performance of SMAJudge. All the experiments were implemented with Python and run on a server with Intel Xeon CPU/64G RAM/NVIDIA RTX 1080Ti.

\subsection{Data}

We collected our experimental dataset from China Judgment Online\footnote{https://wenshu.court.gov.cn}. The dataset contains appeal judgment documents of 33238 criminal appeal cases, where the number of affirmed cases is 27584. These appeal judgment documents cover 618 charges and 15756 law articles. Following \cite{22}, we only kept the law articles occurring more than 100 times. We extracted ${f^l}$, ${r^l}$, ${g}$, ${f^{a}}$, and ${r^{a}}$  from each appeal judgment document using regular expressions. In the following experiments, we used 70\% of the appeal judgment documents as training samples, 10\% as validation samples, and 10\% as testing samples. Table \ref{tab:statistics} lists the statistical information of the experimental dataset.

\begin{table}[htbp]
\footnotesize
\caption{Experimental Dataset Statistics}
\begin{center}
\label{tab:statistics}
\tabcolsep 25pt 
\begin{tabular}{cc}
\toprule
  Statistcs & Numbers  \\\hline
 Total Number of Charges&6 \\
  Total Number of Law Articles&68 \\
  Number of intervals of terms of Penalty&11 \\
 Total Number of Appeal Cases&33238 \\
 Total Number of Affirmed Cases &27584 \\
\bottomrule
\end{tabular}
\end{center}
\end{table}

\subsection{Experimental Settings}
\label{Sec:ExpSetting}
Since the appeal judgment documents in the experimental dataset are in Chinese, we first performed word segmentation on the extracted textual content with Baidu Lexical Analysis of Chinese\footnote{https://github.com/baidu/lac}. Then we employed the word2vec model to embed each word as a 200-dimensional vector. The total number of words is 1.35 million. We set the dimension of BiLSTMs' hidden layers in SMAJudge as 256. In the training process of SMAJudge, the batch size, the learning rate, and the dropout were set as 50, 0.003, and 0.5, respectively.

We used Accuracy (Acc.), Macro Precision (MP), Macro Recall (MR), and Macro $\text{F}_1$ (F1) as the measures of performance. Let $K$ refer to the number of category labels in a classification task, Acc., MP, MR, and F1 are computed as follows:

\begin{equation}
Acc.=\frac{\sum_{i\in [1,K]}TP_{i}+\sum_{i\in [1,K]}TN_{i}}{\sum_{i\in [1,K]}(TP_{i}+TN_{i}+FP_{i}+FN_{i})},
\end{equation}
\begin{equation}
MP=\frac{\sum_{i\in [1,K]}TP_{i}}{\sum_{i\in [1,K]}TP_{i}+\sum_{i\in [1,K]}FP_{i}},
\end{equation}
\begin{equation}
MR=\frac{\sum_{i\in [1,K]}TP_{i}}{\sum_{i\in [1,K]}TP_{i}+\sum_{i\in [1,K]}FN_{i}},
\end{equation}
\begin{equation}
F1=\frac{2\times MP\times MR}{MP+MR}.
\end{equation}
where $TP_i$, $FP_i$, $TN_i$, and $FN_i$ denote the numbers of true positive, false positive, true negative, and false negative samples regarding the $i$-th category, respectively.

\subsection{Performance Comparison on Prediction of $y^a_r$}

We compared the performance of SMAJudge with those of the following six baseline methods. Since all the baselines are designed for a lower court trial, we took $[f^l;f^a]$ as the input of these baselines and set $y^a_r$ as their prediction objective (for single-task-based methods) or one of their prediction objectives (for multi-task-based methods).
\begin{itemize}
	\item \textbf{TFIDF+SVM}\cite{26}: This model extracts a vector representation of the input using TF-IDF features and predicts the appeal judgment with an SVM classifier.
	\item \textbf{TextCNN}\cite{27}: This model encodes the input with a CNN to extract its semantic representation. Then the representation is passed into a full-connection layer followed by a softmax layer to get the probability distribution of the appeal judgment.
	\item \textbf{Seq2Seq}\cite{28}: This model utilizes LSTMs to encode and decode the input. Then it passes the learned representation of the input into a fully-connected layer followed by a softmax layer to get the probability distribution of the appeal judgment.
	\item \textbf{HAN}\cite{29}: The model employs BiGRUs to obtain the word-level and sentence-level embeddings of the input with an attention mechanism. Then the embeddings are passed into a Multi-Layer Perception followed by a full-connection layer. Finally, a sigmoid classifier outputs the prediction of the appeal judgment.
	\item \textbf{TOPJUDGE}\cite{22}: We modified the topological multi-task learning framework proposed in \cite{22} by adding the prediction of $y^a_r$ into its subtask list. 
	\item \textbf{BERT}: We also implemented a variation of SMAJudge as a baseline, which has the same architecture with SMAJudge but replaces the BiLSTM encoders with BERT.
\end{itemize}

In the training processes of TextCNN, Seq2Seq, HAN, TOPJUDGE, and BERT, the settings of the embedding dimension, the learning rate, the dropout, and the batch size were the same with SMAJudge.

\begin{table}[htbp]
\footnotesize
\caption{Performance Comparisons with Baseline Methods on Predicting the Appellate Court's Ruling on the Lower Court's Judgment}
\begin{center}
\begin{tabular}{c|c|cccc}
\toprule
  \rule{0pt}{10pt}  & \textbf{Type}& \textbf{Acc}.& \textbf{MP}& \textbf{MR}& \textbf{F1}\\
\hline
  \rule{0pt}{10pt}TFIDF+SVM&& 85.0&79.0&55.0&56.0\\
\rule{0pt}{10pt}TextCNN&Single task&87.1&80.6&64.4&68.1 \\
\rule{0pt}{10pt}Seq2Seq&&88.0&83.9&64.2&68.3 \\
\rule{0pt}{10pt}HAN&&87.5&84.2&64.3& 68.4 \\
\hline
\rule{0pt}{10pt}TOPJUDGE&&\textbf{88.2}&\textbf{85.2}&66.5&70.9 \\
\rule{0pt}{10pt}BERT&Multi-task&86.0&75.2&66.2&69.0 \\
\rule{0pt}{10pt}\textbf{SMAJudge}&&86.7&75.4&\textbf{72.2}&\textbf{73.7} \\
\bottomrule
\end{tabular}
\label{table:comp_res}
\end{center}
\end{table}

Table \ref{table:comp_res} summarizes the experimental results obtained by all compared methods. We derive the following observations from the results. 
\begin{itemize}
\item \emph{The performance of deep-learning-based methods is better than that of the shallow machine learning model (i.e., TFIDF+SVM).} We can see that TFIDF+SVM performs the worst in terms of all measures. This observation is consistent with most existing studies on other LJP tasks. It indicates that deep learning models such as CNN, LSTM, GRU are better in extracting the latent semantics in the description of facts, which can improve the results of AJP. 
\item \emph{Multitask-learning-based methods' overall performance is better than that of single-task-based methods.} Although the accuracy and MP scores of BERT and SMAJudge are lower than those of TextCNN, Seq2Seq, and HAN, they obtain higher MR and F1 scores. The results indicate that utilizing the correlation among relevant subtasks can also improve AJP performance. 
\item \emph{SMAJudge obtains the best overall performance among all the compared methods.} Although the accuracy and MP scores of SMAJudge are worse than those of many of the baselines, it obtains the highest MR and F1 scores. That is, SMAJudge's overall performance is the best. The relative improvement of SMAJudge over the best baseline method (i.e., TOPJUDGE) in terms of MR and F1 are 8.57\% and 3.95\%, respectively. Since there is an imbalance between samples of affirmed cases and those of other cases, if a model determines most of the test samples as ``affirmed'' (i.e., $\hat{y}^a_r=0$), it may have high accuracy and MP scores. However, its performance in terms of MR will be terrible, as it classifies more samples of ``unaffirmed'' case as ``affirmed''. SMAJudge obtains a good MR on the small class, making its F1 score the best among all baselines. 
\end{itemize}

In summary, the experimental results indicate that by modeling the complete proceedings logic from a lower court to an appellate court properly, SMAJudge obtains better overall performance than all the baselines.

\subsection{Performance Comparisons on Other Subtasks}

To validate the effectiveness of SMAJudge on the other subtasks in $M^l$ and $M^a$, we compared its performance on $t_1^l$, $t_2^l$, $t_3^l$, and $t_{art}^a$ with those of the other two multi-task-based baseline methods, i.e., TOPJUDGE and BERT. Table \ref{tab:subtasks} gives the comparison results, from which we have the following observations.

\begin{table*}[htbp]
\footnotesize
\renewcommand{\arraystretch}{1.5}
\caption{Performance Comparisons with Multi-Task-Based Baseline Methods on Four Subtasks}
\begin{center}
\tabcolsep 4pt
\begin{tabular}{c|cccc|cccc|cccc|cccc}
\toprule
Subtasks& \multicolumn{4}{c|}{$t^l_1$ in $M^l$} & \multicolumn{4}{c|}{$t^l_2$ in $M^l$}&  \multicolumn{4}{c|}{$t^l_3$ in $M^l$}& \multicolumn{4}{c}{$t^a_{l}$ in $M^a$}\\
\hline
Metrics&Acc.&MP&MR&F1&Acc.&MP&MR&F1&Acc.&MP&MR&F1&Acc.&MP&MR&F1\\
\hline
TOPJUDGE&86.0&76.3&69.1&70.9&87.3&\textbf{80.1}&73.2&74.4&37.5&\textbf{41.4}&36.4&34.9&85.9&73.3&37.8&43.4 \\
BERT&86.9&76.2&\textbf{74.2}&73.1&\textbf{87.4}&78.0&\textbf{77.7}&\textbf{77.0}&\textbf{38.5}&38.9&\textbf{39.0}&\textbf{38.3}&86.3&63.7&\textbf{47.1}&50.1 \\
\textbf{SMAJudge}&\textbf{87.2}&\textbf{77.2}&73.6&\textbf{73.9}&87.0&79.9&75.4&76.1&36.9&40.2&36.6&37.5&\textbf{87.0}&\textbf{78.6}&45.1&\textbf{51.0} \\

\bottomrule
\end{tabular}
\label{tab:subtasks}
\end{center}
\end{table*}

\begin{itemize}
	\item \emph{Performance comparison with TOPJUDGE.} Although the accuracy and MP scores of SMAJudge are worse than TOPJUGDE on subtask $t^l_2$ and $t^l_3$, SMAJudge outperforms TOPJUDGE in terms of MR and F1 on all subtasks. The results indicate that SMAJudge has a better overall performance on all subtasks than TOPJUDGE.
	\item \emph{Performance comparison with BERT.} The overall performance of SMAJudge in terms of F1 is better than BERT on the two relevant law article prediction subtasks, i.e., $t^l_1$ and $t^a_{law}$, but is worse than BERT on $t^l_2$ and $t^l_3$. Note that BERT has the same architecture as SMAJudge except the encoder of facts. The results may be because determining which ones are relevant from dozens of law articles needs more fact details than the other two subtasks do. However, the input length limit causes it to lose details of the facts.
\end{itemize}

In summary, the overall performances of SMAJudge and BERT according to F1 are better than TOPJUDGE. As the SMAJudge and BERT have the same architecture, the results indicate that the design of sequential topological multi-task learning can further jointly promote the performance on all subtasks compared with directly taking all subtasks as parallel in a single multi-task learning framework.

\subsection{Sensitivity to the Scale of Training Data}

\begin{figure}[htbp]
\centering
\subfloat[]{\includegraphics[width=2.1in]{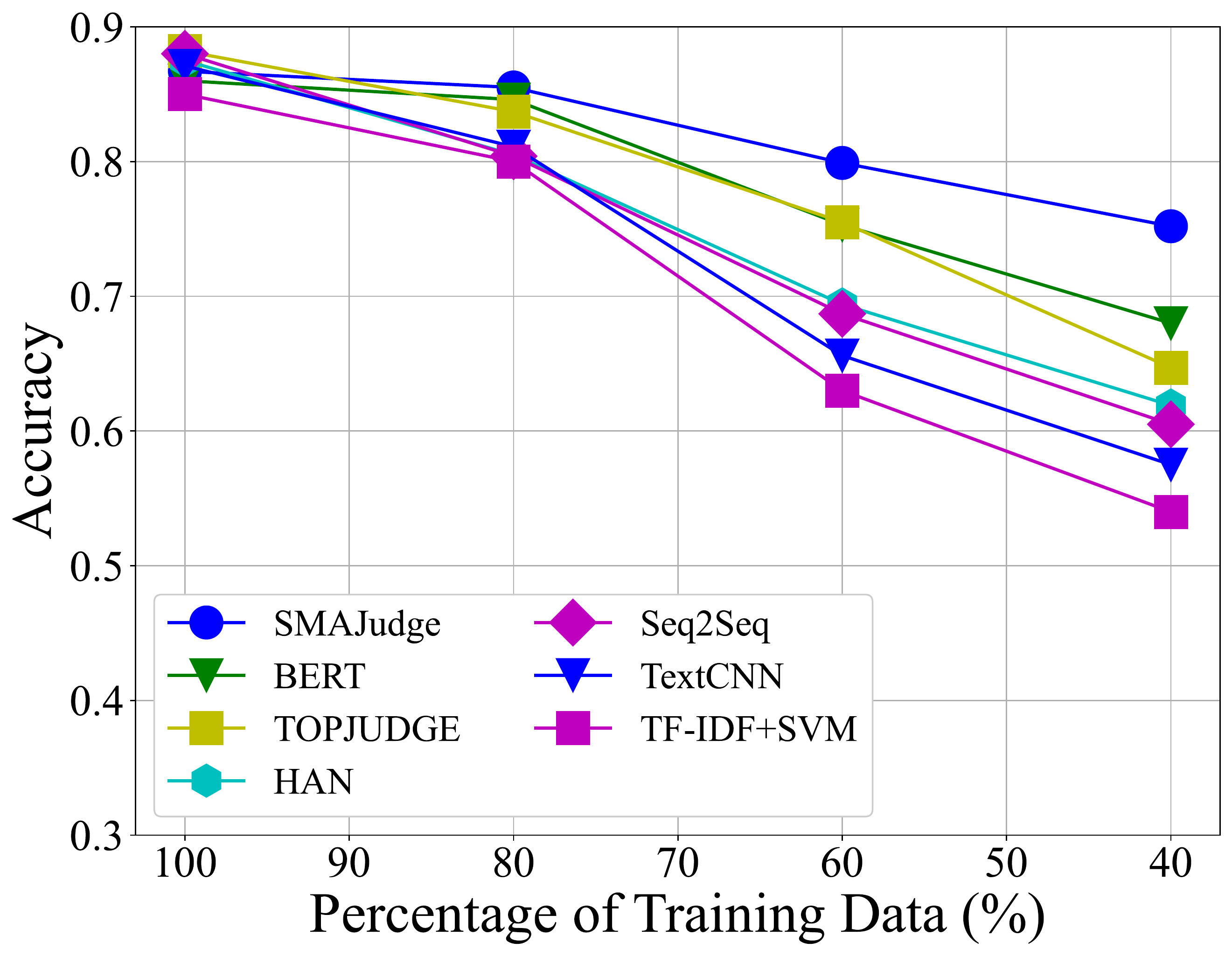}%
\label{}}
\hfil
\subfloat[]{\includegraphics[width=2.1in]{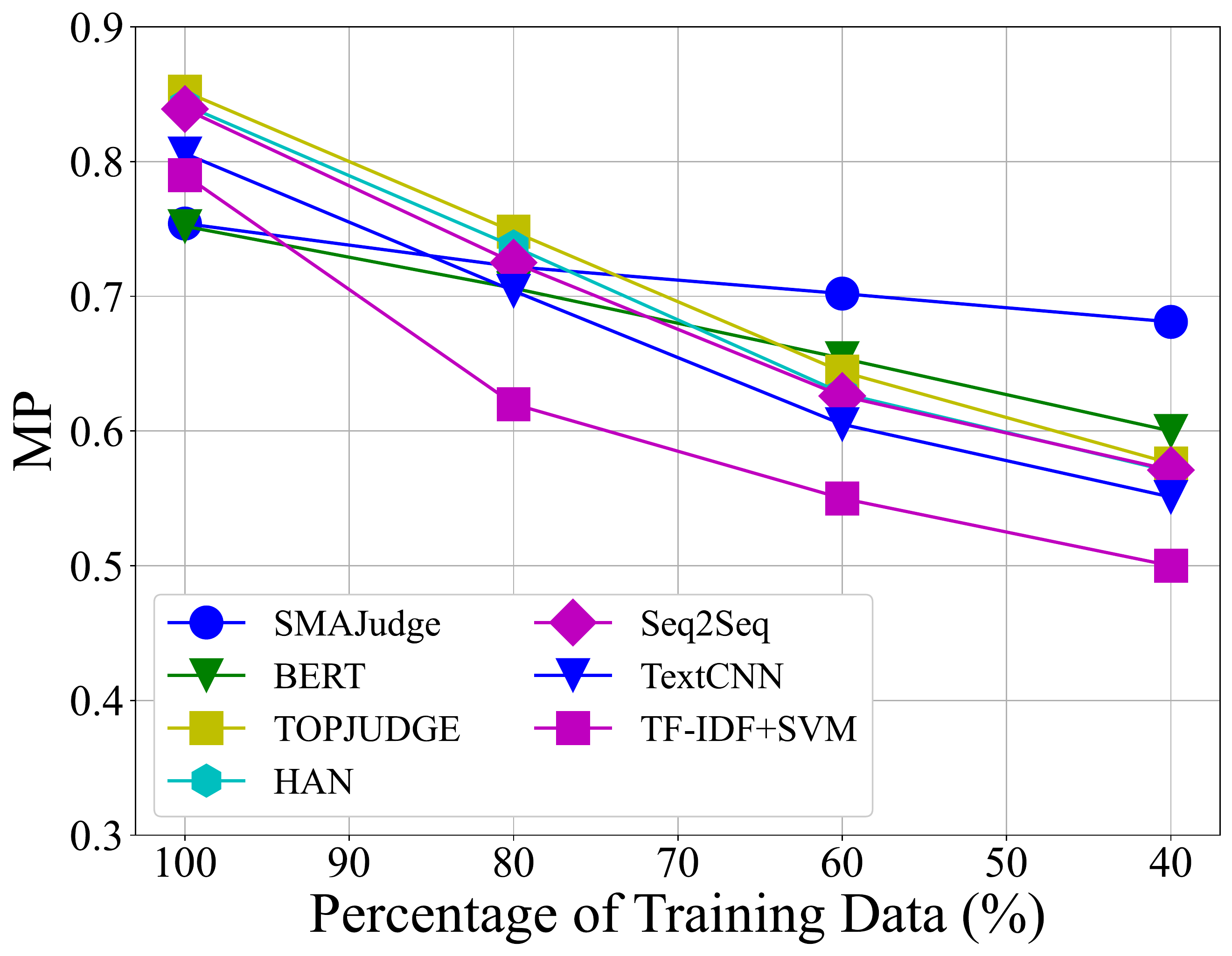}%
\label{}}
\hfil
\subfloat[]{\includegraphics[width=2.1in]{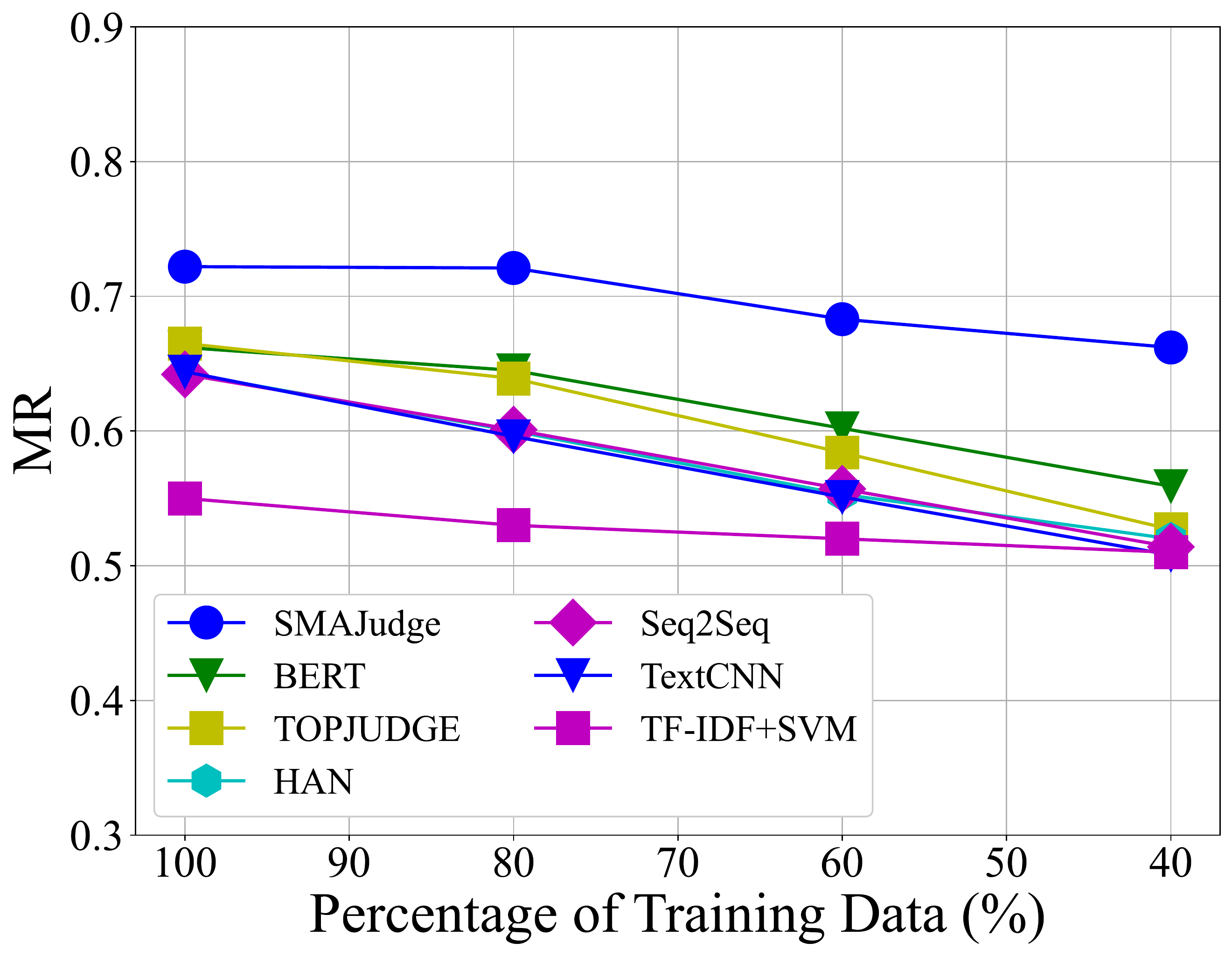}%
\label{}}
\hfil
\subfloat[]{\includegraphics[width=2.1in]{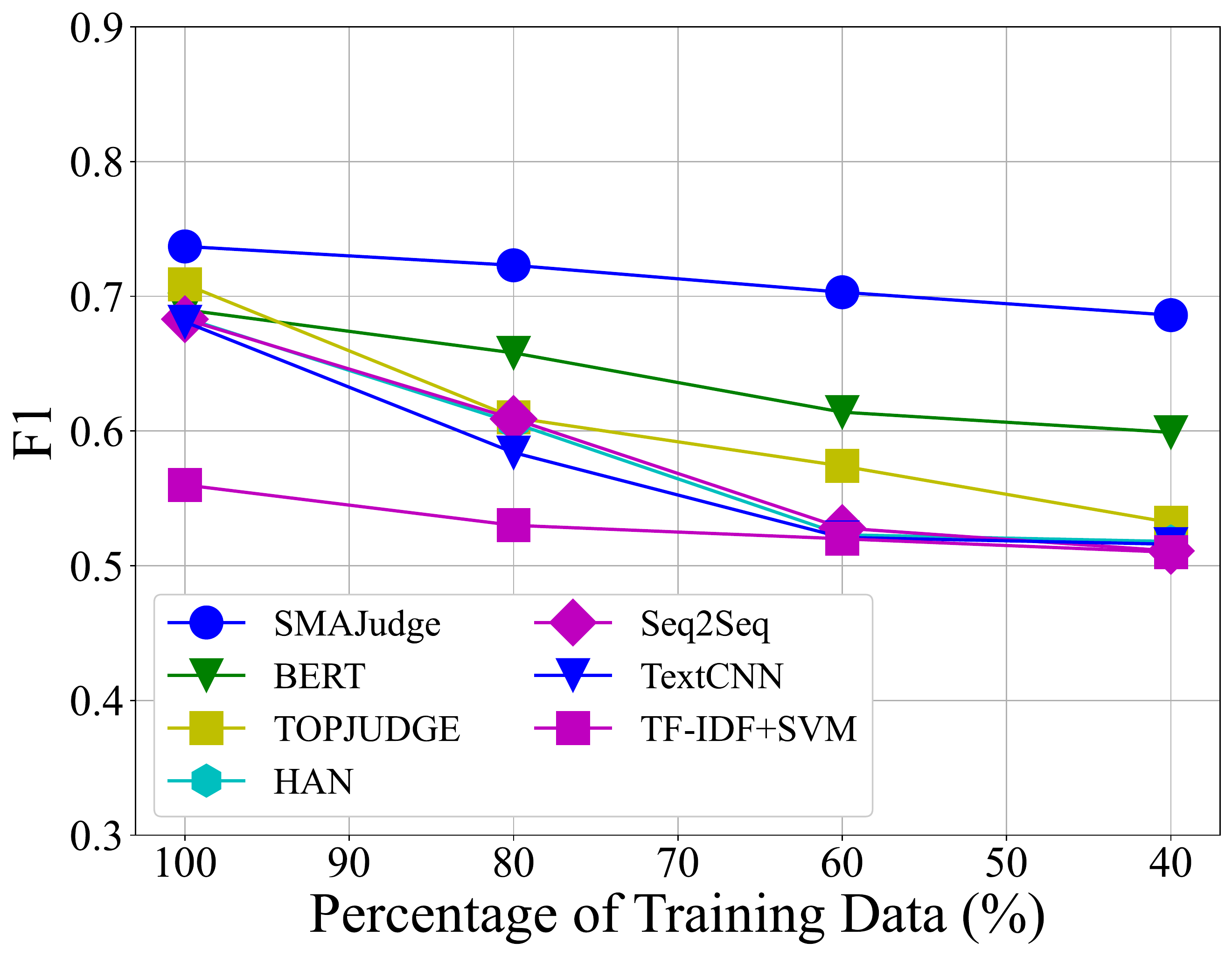}%
\label{}}
\caption{Performance variations of SMAJudge and baseline methods in terms of Acc. (a), MP (b), MR (c), and F1 (d) with the size of training data reducing.}
\label{fig:sizered}
\end{figure}

We also examined the sensitivity of SMAJudge and other baseline methods to the scale of training data. We trained all the methods listed in Table \ref{table:comp_res} with 100\%, 80\%, 60\%, and 40\% of the training data, respectively. Figure \ref{fig:sizered} shows how the performance of all methods varies with the reduction of training data. 

From the results, we can see that the performances of all methods decrease with the reduction of training data. However, the performance of SMAJudge is less sensitive to the scale of training data compared with most baseline methods. For instance, when reducing the scale of training data from 100\% to 40\%, the F1 scores of BERT, TOPJUDGE, HAN, Seq2Seq, and TextCNN decreased by 9.1\%$\sim$17.7\%, while the F1 score of SMAJudge decreased by only 5.1\% (see Figure \ref{fig:sizered}d). The F1 of TF-IDF+SVM is not sensitive to reducing training data either but is always worse than that of SMAJudge.  We can observe similar phenomena from the performance variation in accuracy, MP, and MR  (see Figures \ref{fig:sizered}a, \ref{fig:sizered}b, and \ref{fig:sizered}c). 

Although SMAJudge performs worse than some baseline methods in terms of accuracy and MP when training with 100\% training data, its performance is significantly better than all baseline methods when training with 60\% and 40\% of the data. Notably, when training with 40\% of the training samples, the relative improvements of SMAJudge over the best baseline methods in terms of accuracy, MP, MR, and F1 are 10.59\%, 13.5\%, 18.43\%, and 14.52\%, respectively.

In summary, the results reveal that SMAJudge can obtain much better performance with a small scale of training data compared with baseline methods.

\subsection{Ablation Study}

We conducted ablation tests on components in SMAJudge to evaluate their significance and necessity. Table \ref{table:ablation} lists the prediction results on $y^a_r$, where the designs of the models are as follows:
\begin{itemize}
\item \textbf{M$^l$M$^a$}: To build this model, we first trained $M^l$ and $M^a$ separately and obtained the vectors $\mathbf{h}^l$ and $\mathbf{h}^a$ representing the lower court and the appellate court's understanding of case facts, respectively. Then we computed the similarity between $\mathbf{h}^l$ and $\mathbf{h}^a$ to predict the appeal judgment. If their similarity is larger than 0.5, the model outputs $\hat{y}^a_r=0$. Otherwise, the model outputs $\hat{y}^a_r=1$.

\item \textbf{SMAJudge$_{-att}$}: To build this model, we removed the attention mechanism in SMAJudge and only used $f^a$ as the input of $M^a$ in AJPSML instead of using $[f^a;g]$.
\item \textbf{SMAJudge$_{-dep}$}: To build this model, we removed the dependencies between subtasks in SMAJudge and considered the relationships among all the subtasks as parallel.
\end{itemize}

\begin{table}[htbp]
\footnotesize
\caption{Results of ablation tests}
\begin{center}
\tabcolsep 15pt
\begin{tabular}{c|cccc}
\toprule
  \rule{0pt}{10pt}  & \textbf{Acc}.& \textbf{MP}& \textbf{MR}& \textbf{F1}\\
\hline
  \rule{0pt}{10pt}M$^{a}$M$^{l}$&65.7&56.5&64.0&57.9\\
   \rule{0pt}{10pt}SMAJudge$_{-att}$&86.5&75.2&70.6&72.5 \\
 \rule{0pt}{10pt}SMAJudge$_{-dep}$&85.7&72.3&70.0&71.0 \\
\rule{0pt}{10pt}\textbf{SMAJudge}&\textbf{86.7}&\textbf{75.4}&\textbf{72.2}&\textbf{73.7} \\
\bottomrule

\end{tabular}
\label{table:ablation}
\end{center}
\end{table}

From Table \ref{table:ablation}, we can see that the performance of SMAJudge is much better than that of $\text{M}^l\text{M}^a$ in terms of all measures, which indicates that the strategy of jointly modeling the dependency between the two judgment proceedings with sequential multitask learning is more suitable to simulate the decision logic of the appellate court. SMAJudge also outperforms $\text{SMAJudge}_{-att}$ in terms of all measures. We found that the attention mechanism in SMAJudge can guide the model to focus on the parts of facts having significant impacts on the appellate court's decision, which makes SMAJudge have better prediction results. Similarly, the comparison between the results of $\text{SMAJudge}_{-dep}$ and SMAJudge shows that modeling the dependencies between subtasks can also improve the performance of predicting $y^a_r$.

In summary, the ablation tests demonstrate that all the critical components in SMAJudge contribute to its overall performance.

\subsection{Interpretability Analysis of SMAJudge}

\begin{figure}[htbp]
\centerline{\includegraphics[width=15cm]{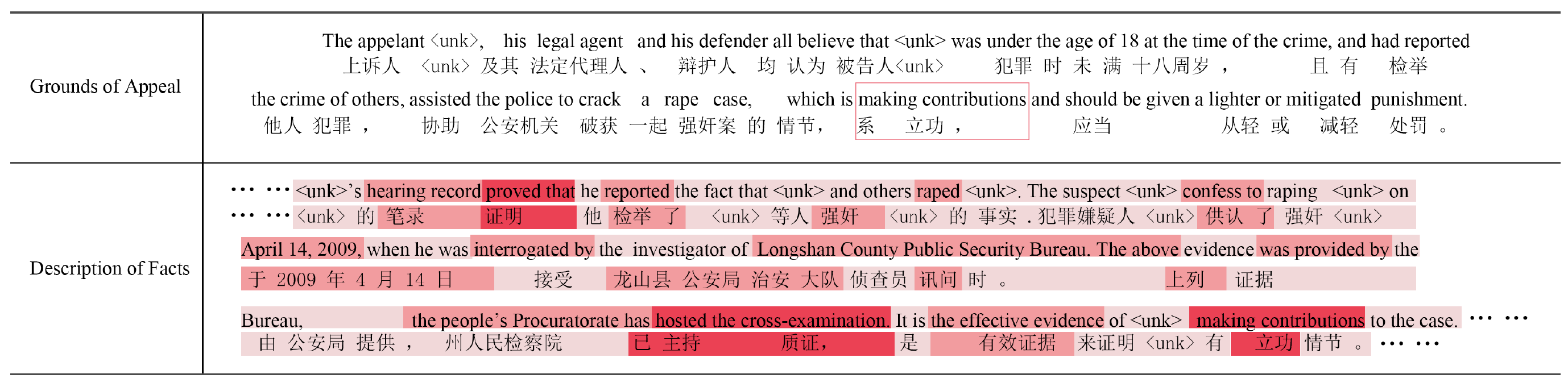}}
\caption{Visualization of Attention Mechanism on Case I}
\label{fig:case1}
\end{figure}

\begin{figure*}[htbp]
\centerline{\includegraphics[width=15cm]{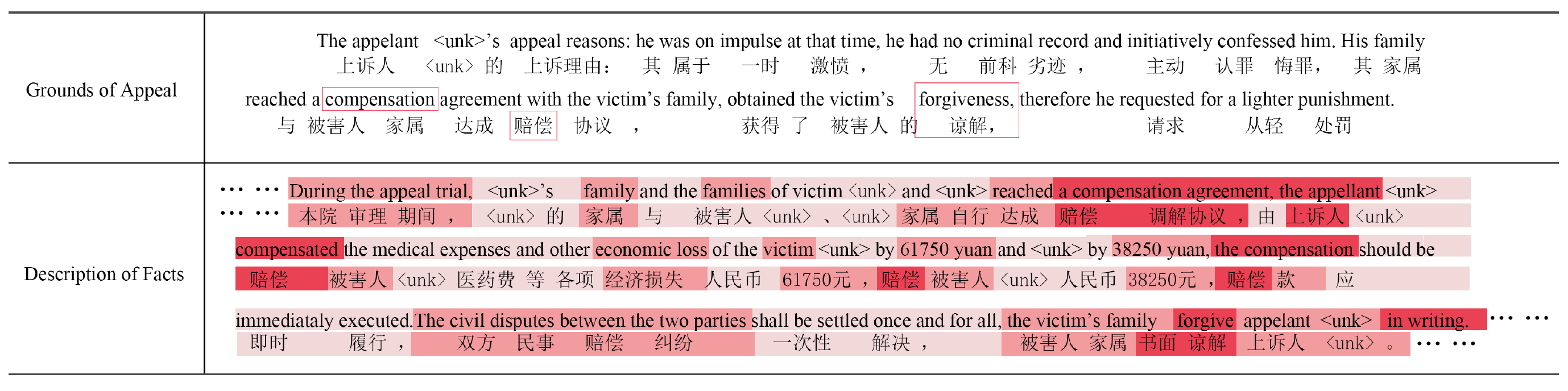}}
\caption{Visualization of Attention Mechanism on Case III}
\label{fig:case3}
\end{figure*}

We visualized the results of the attention mechanism in $\text{M}^a$ to evaluate SMAJudge's interpretability intuitively. Figure \ref{fig:case1}$\sim$\ref{fig:case3} shows the visualization results of three appeal cases, where the appellate courts overturned the lower courts' judgments. The visualization of each case enumerates the grounds of appeal and the description of case facts. The colors of words in the description of facts reflect their attention weights in the corresponding encoder, i.e., darker words get higher attention weights in the encoding process, and their information is more critical to the encoder.

\begin{figure*}[htbp]
\centerline{\includegraphics[width=15cm]{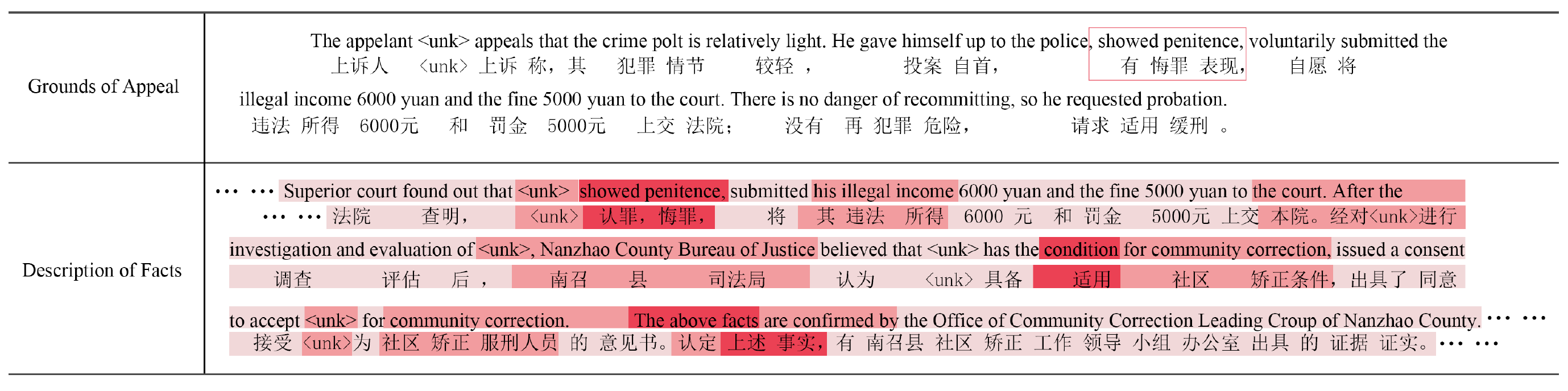}}
\caption{Visualization of Attention Mechanism on Case II}
\label{fig:case2}
\end{figure*}

In the visualizations of all these three cases,  critical words or phrases indicating why appellate courts overturn lower courts' judgment always get higher attention weights than others. In Case-I, ``making contributions'' is one of the phrases that obtain the highest attention score. It corresponds to the fact that the appellant ``reported'' other individuals' ``rape'' behaviors, and the reported crime has been examined as ``effective evidence.'' This fact implies why the appellate court changed the lower court's judgment. The words ``reported,'' ``rape,'' and ``effective evidence'' also get high attention sores. In Case-II, the appellate court changed the lower court's judgment because the appellant ``showed penitence'' and submitted ``his illegal income'' ``to the court.'' Therefore, the appellant had the ``condition'' ``for community correction.'' We can see that all critical words and phrases are highlighted with high attention scores. Similarly, in Case-III, the words ``compensation,'' ``compensated,'' and ``forgive'' obtain the highest attention score, which indicates the factors making the appellate court reduced the appellant's penalty.

In summary, with the attention mechanism, the encoding procedure of $\text{M}^a$ can capture the vital information in the description of facts that leads to the ruling of the appellate court over the lower court's judgment. For ``overturned'' cases, the words and phrases with high attention scores can implicitly explain why the appellate court changed the lower court's decision, which promotes the interpretability of SMAJudge.

\section{Conclusion and Future Work}
\label{sec:conclusions}
This paper proposes a sequential topological multitask learning framework named SMAJudge for solving the AJP task. SMAJudge uses the textual description of an appeal case as input, including case facts, the judgment of a lower court, and grounds of appeal. It employs two sequential multitask learning components to simulate the decision logic of an appellate court and uses an attention mechanism to improve the interpretability of prediction. Experimental results on a real-world dataset show that SMAJudge has a better overall performance than all baseline methods. Moreover, SMAJudge can obtain significantly better performance with a small training data scale than baseline methods. Ablation tests demonstrate the significance and necessity of all the critical components in SMAJudge. We also prove that SMAJudge can produce prediction results with good interpretability benefiting from the attention mechanism.

In the future, we will focus on two directions. First, we will add more subtasks (e.g., court view generation) into SMAJudge and explore better model architectures. Second, we will study better ways to mitigate the impact of data imbalance.

\Acknowledgements{This work was supported in part by the National Key Research and Development Program of China under Grant 2018YFC0830100 and Grant 2018YFC0830102, in part by the National Natural Science Foundation of China under Grant 61602281,  in part by the Shandong Provincial Key Research and Development Program under Grant 2019JZZY020129, Grant 2019JZZY010134, and Grant 2019JZZY010132, and in part by the Shandong Provincial Natural Science Foundation of China under Grant ZR2020KF035.}

\end{document}